\pdfoutput=1

\documentclass[11pt]{article}

\usepackage{naacl2021}

\usepackage{times}
\usepackage{latexsym}

\usepackage{caption}
\usepackage{subcaption}

\usepackage{algorithm}
\usepackage[noend]{algpseudocode}

\usepackage[T1]{fontenc}
\usepackage[utf8]{inputenc}

\usepackage{microtype}
\newif\ifcomment\commenttrue

\usepackage[a-1b]{pdfx}

\usepackage{framed}
\usepackage{mdwlist}
\usepackage{latexsym}
\usepackage{colortbl}
\usepackage{xcolor}
\usepackage{nicefrac}
\usepackage{booktabs}
\usepackage{amsfonts}
\usepackage[T1]{fontenc}
\usepackage{bold-extra}
\usepackage{amsmath}
\usepackage{amssymb}
\usepackage{bm}
\usepackage{graphicx}
\usepackage{mathtools}
\usepackage{microtype}
\usepackage{multirow}
\usepackage{multicol}
\usepackage{latexsym,comment}
\usepackage[normalem]{ulem}

\newcommand*{\missingreference}{{\Huge \colorbox{red}{?reference?}}}
\newcommand*{\missingcitation}{{\Huge \colorbox{red}{?citation?}}}
\makeatletter
\def\@setref#1#2#3{%
  \ifx#1\relax
    \protect\G@refundefinedtrue
    \nfss@text{\reset@font\missingreference}%
    \@latex@warning{Reference `#3' on page \thepage \space
      undefined}%
  \else
    \expandafter#2#1\null
  \fi}
\def\@citex[#1]#2{\leavevmode
  \let\@citea\@empty
  \@cite{\@for\@citeb:=#2\do
    {\@citea\def\@citea{,\penalty\@m\ }%
      \edef\@citeb{\expandafter\@firstofone\@citeb\@empty}%
      \if@filesw\immediate\write\@auxout{\string\citation{\@citeb}}\fi
      \@ifundefined{b@\@citeb}{\hbox{\reset@font\missingcitation}%
        \G@refundefinedtrue
        \@latex@warning
        {Citation `\@citeb' on page \thepage \space undefined}}%
      {\@cite@ofmt{\csname b@\@citeb\endcsname}}}}{#1}}
\makeatother

\newcommand{\gem}[1]{\mbox{\textsc{gem}}}
\newcommand{\abr}[1]{\textsc{#1}}

\newcommand{\hidetext}[1]{}
\newcommand{\ignore}[1]{}

\ifcomment
  \newcommand{\pinaforecomment}[3]{\colorbox{#1}{\parbox{.8\linewidth}{#2: #3}}}
\else
  \newcommand{\pinaforecomment}[3]{}
\fi

\newcommand{\smallurl}[1]{ \begin{tiny}\url{#1}\end{tiny}}

\definecolor{lightblue}{HTML}{3cc7ea}
\definecolor{CUgold}{HTML}{CFB87C}
\definecolor{grey}{rgb}{0.95,0.95,0.95}
\definecolor{ceil}{rgb}{0.57, 0.63, 0.81}
\definecolor{UMDred}{HTML}{ed1c24}
\definecolor{UMDyellow}{HTML}{ffc20e}

\newcommand{\splash} {\abr{splash}}
\newcommand{\spider} {\abr{spider}}
\newcommand{\modelname} {\abr{nl-edit}}

\title{NL-EDIT:\\ Correcting Semantic Parse Errors through Natural Language Interaction}

\author{
  Ahmed Elgohary$^\bigstar$\thanks{\hspace{1.5mm}Most of the work was done while the first author was an intern at Microsoft Research.}, 
  Christopher Meek$^\clubsuit$, Matthew Richardson$^\clubsuit$, Adam Fourney$^\clubsuit$,\\
  {\bf Gonzalo Ramos$^\clubsuit$, Ahmed Hassan Awadallah$^\clubsuit$}\\
  $^\bigstar$University of Maryland \hspace{1cm}
  $^\clubsuit$Microsoft Research \\
  \small \texttt{elgohary@cs.umd.edu} \hspace{0.2cm} 
  \texttt{\{meek,mattri,adamfo,goramos,hassanam\}@microsoft.com} 
}
\begin{document}
\maketitle
\begin{abstract}
 We study semantic parsing in an interactive setting
 in which users correct errors with natural language feedback.
 We present \modelname, a model for interpreting natural language feedback in the interaction context to generate a sequence of edits that can be applied to the initial parse to correct its errors. 
 We show that \modelname\ can boost the accuracy of existing text-to-SQL parsers by up to 20\% with only one turn of correction. We analyze the limitations of the model and discuss directions for improvement and evaluation. The code and datasets used in this paper
 are publicly available at \url{http://aka.ms/NLEdit}.

\end{abstract}

\section{Introduction\label{sec:intro}}
\begin{figure}[!t]
  \centering
  \includegraphics[width=\columnwidth]{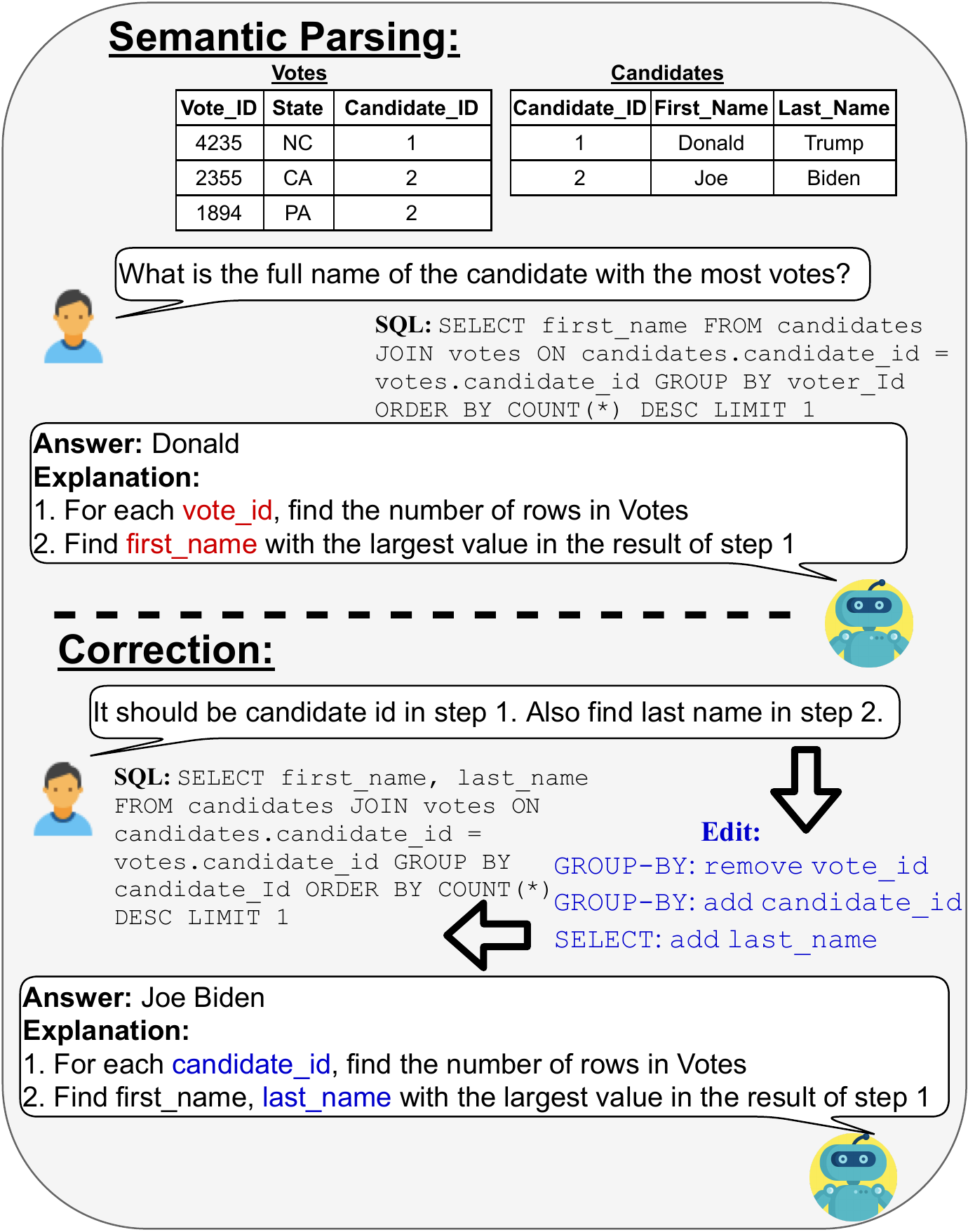}
    \caption{Example human interaction with \modelname\
    to \textit{correct} an initial parse through natural language feedback. In the  \textbf{Semantic Parsing Phase} (top),
    an off-the-shelf parser generates an initial SQL query
    and provides an answer paired with an \textit{explanation} of
    the generated SQL. In the \textbf{Correction Phase} (bottom), the user reviews the
    explanation and provides \textit{feedback} that describes how
    the explanation should be corrected. The system parses the feedback
    as a set of \textit{edits} that are applied to the initial parse 
    to generate a corrected SQL.
    }
  \label{fig:example}
 \end{figure}

Major progress in natural language processing has been made towards fully automating challenging tasks such as question answering, translation, and summarization. On the other hand, several studies have argued that machine
learning systems that can explain their own predictions ~\cite{doshi2017towards} and learn interactively from
their end-users ~\cite{amershi2014power} can result in better user experiences and more effective learning systems. 
We develop \modelname---an approach that employs both explanations and interaction in the context of semantic
parsing.

Most existing systems frame semantic parsing as a one-shot 
translation from a natural language question to the corresponding logical form (e.g., SQL query)~\cite[inter alia]{yu2018syntaxsqlnet,guo-etal-2019-towards,rat-sql-2020}. A growing body of recent work 
demonstrates that semantic parsing systems can be improved by including users in the parsing loop---giving
them the affordance to examine the parses, judge their correctness, and provide feedback accordingly. The feedback often comes in the form of a binary correct/incorrect signal~\cite{iyer-etal-2017-learning}, answers to a multiple-choice
question posed by the system~\cite{gur-etal-2018-dialsql,yao-etal-2019-model},  or suggestions of edits that can be applied to the parse~\cite{su2018natural}.

Unlike other frameworks for interactive semantic parsing that typically expect users to judge the correctness of the execution result or induced logical form, \newcite{Elgohary2020Speak} introduced a framework
for interactive text-to-SQL in which induced SQL 
queries are fully explained in natural language to users, who in turn, 
can correct such parses through natural language feedback~(Figure~\ref{fig:example}).
They construct the  \splash\ dataset and use it to evaluate baselines for the
semantic parse correction with natural language feedback task they introduce.

We present a detailed analysis of the feedback and the differences between the initial (incorrect) and the correct parse. We argue that a correction model should be able to interpret the feedback in the context of other elements of the interaction (the original question, the schema, and the explanation of the initial parse). We  observe from \splash\ that  most feedback utterances tend to describe a few edits that the user desires to apply to the initial parse. As such, we pose the correction task as a semantic parsing problem that aims to convert  natural language feedback to a sequence of edits that can be deterministically applied to the initial parse to correct it. We use the edit-based modeling framework to show that  we can effectively generate synthetic data to pre-train the correction model leading to clear performance gains.

We make the following contributions: (1) We present a scheme for representing SQL query Edits 
that benefits both the modeling and the analysis of the correction task, 
(2) we present \modelname, an edit-based model for interactive text-to-SQL with natural language feedback. We show that \modelname\ outperforms baselines in~\cite{Elgohary2020Speak} by more than 16 points, (3) We demonstrate that we can generate synthetic data through the edit-based framing and that the model can effectively use this data to improve its accuracy and (4) We present a detailed analysis of the model performance including studying the effect of different components, generalization to errors of state-of-the-art parsers, and outline directions for future research.

\section{Background \label{sec:background}}
In the task of text-to-SQL parsing, the objective is given a database schema (tables, columns, and primary-foreign key relations) and a natural language question, generate a SQL query that answers the question
when executed against the database. 
Several recent text-to-SQL models have been introduced \cite[inter alia]{yu2018syntaxsqlnet, zhang19, guo-etal-2019-towards, rat-sql-2020} as a result of the availability of \spider\ \cite{yu-etal-2018-spider},
a large dataset of schema, questions and gold parses spanning several databases
in different domains.

The task of SQL parse correction with natural language feedback~\cite{Elgohary2020Speak} aims to correct an erroneous parse based on  natural language feedback collected from the user. Given a question, a database schema, an
incorrect initial parse, natural language feedback on the initial parse, the task is to generate a corrected parse.

To study this problem, \newcite{Elgohary2020Speak} introduced the \splash\ dataset. \splash\ was created by showing annotators questions and a natural language explanation of incorrect parses and asking them to provide feedback, in natural language, to correct the parse. The dataset contained 9,314 question-feedback pairs. Like the \spider\ dataset, it was split into train-dev-test sets by database to encourage the models to generalize to new unseen databases. They contrast the task with conversational semantic parsing~\cite{suhr-etal-2018-learning, Yuetal19Sparc, yu-etal-2019-cosql, Andreas2020Task} 
and show that the two tasks are distinct and are addressing different aspects of utilizing context. They establish several baseline models  and show that the task is challenging for state-of-the-art semantic parsing models. We use these as baselines for this work.

\section{SQL Edits\label{sec:sqledits}}
\begin{figure*}[!t]
	\centering
	\includegraphics[width=\linewidth]{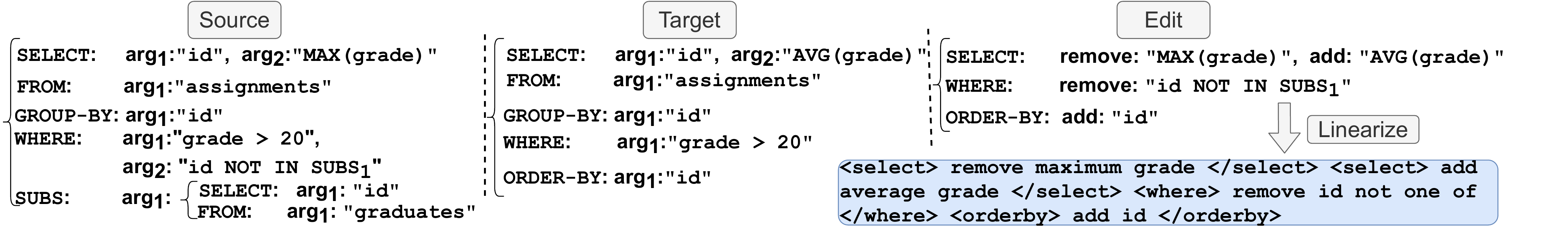}
	\caption{Edit for transforming the \textbf{source} query 
\texttt{``SELECT id, MAX(grade) FROM assignments WHERE grade > 20 
AND id NOT IN (SELECT id from graduates) GROUP BY id''}~to~the
\textbf{target} \texttt{``SELECT id, AVG(grade) FROM assignment WHERE
grade > 20 GROUP BY id ORDER BY id''}. The source and target are
represented as sets of clauses (left and middle). The set of 
\textbf{edits} and its \textbf{linearized} form~(Section~\ref{sec:model}) are shown on the right. Removing the condition \texttt{``id NOT IN \texttt{$\text{SUBS}_1$}''}
makes the subquery unreferenced, hence pruned from the edit.}
\label{fig:queryview}
\vskip -1em
\end{figure*}

We define a scheme for representing the edits required to transform one SQL
query to another. We use that scheme both in our model and analysis.
Our goal is to balance the granularity of the edits---too fine-grained 
edits result in complex structures that are challenging for models to learn,
and too coarse-grained edits result in less compact structures that are harder for
models to generate.

We view a SQL query as a set of clauses~(e.g, \texttt{SELECT}, \texttt{FROM}, \texttt{WHERE}), each clause has a 
sequence of arguments~(Figure~\ref{fig:queryview}). We mirror the SQL clauses
\texttt{SELECT}, \texttt{FROM}, \texttt{WHERE}, \texttt{GROUP-BY},
\texttt{ORDER-BY}, \texttt{HAVING}, and \texttt{LIMIT}. 
For subqueries, we define a clause \texttt{SUBS} whose arguments are recursively defined as sets of
clauses. Subqueries can be linked to the main query in two ways:
either through an \texttt{IEU} clause (mirrors SQL \texttt{INTERSECT/EXCEPT/UNION})
whose first argument is one of the keywords \texttt{INTERSECT}, \texttt{EXCEPT}, \texttt{UNION} and
its second argument is a pointer to a subquery in \texttt{SUBS}.
The second is through nested queries where the arguments of some of the
clauses (e.g., \texttt{WHERE}) can point at subqueries in \texttt{SUBS}
(e.g., \texttt{``id NOT IN $\text{SUBS}_1$''}).

With such view of two queries $\mathcal{P}_{source}$ and $\mathcal{P}_{target}$, 
we define their edit $\mathcal{D}_{source\rightarrow target}$ as the set of 
clause-level edits $\{\mathcal{D}^{c}_{source\rightarrow target}\}$
for all types of clauses $c$ that appear in
$\mathcal{P}_{source}$ or $\mathcal{P}_{target}$~(Figure~\ref{fig:queryview}).
To compare two clauses of type $c$,
we simply exact-match their arguments: unmatched arguments in the
source (e.g., \texttt{MAX(grade)} in \texttt{SELECT}) are added
as to-remove arguments to the corresponding edit clause,  
and unmatched arguments
in the target (e.g., \texttt{``id''} in the \texttt{ORDER-BY}) are
added as to-add arguments.

Our current implementation follows \spider's assumption that the
number of subqueries is at most one which implies that 
computing edits for different clauses can be done independently
even for the clauses that reference a subquery~(e.g., \texttt{WHERE} 
in~Figure~\ref{fig:queryview}). The edit of the \texttt{SUBS} clause
is recursively computed as the edit between two queries~(any of them can be empty); 
the subquery of source and the subquery of target, i.e.,
$\mathcal{D}^{\texttt{SUBS}}_{source\rightarrow target} = \mathcal{D}_{source:\texttt{SUBS}_1\rightarrow target:\texttt{SUBS}_1}$. We keep track of the edits to the arguments that
reference the subquery. After all edit clauses are computed, we prune the 
edits of the \texttt{SUBS} clause if the subquery will no longer
be referenced~(\texttt{$\text{SUBS}_1$} in~Figure~\ref{fig:queryview}). 
We follow the \spider\ evaluation and discard the values in
\texttt{WHERE/HAVING} clauses.

Throughout this paper, we refer to the number of add/remove 
operations in an edit as the \textbf{Edit Size},
and we denote it as
$|\mathcal{D}_{source\rightarrow target}|$. For example,
the edit in Figure~\ref{fig:queryview} is of size four.

\section{Model\label{sec:model}}
We follow the task description in Section~\ref{sec:background}: 
the inputs to the model are the elements of the interaction---question, schema,
an initial parse $\tilde{\mathcal{P}}$, and feedback. The model predicts a corrected 
$\bar{\mathcal{P}}$. The gold parse $\hat{\mathcal{P}}$ is available for training.
Our model is based on integrating two key ideas in an encoder-decoder architecture.
We start with a discussion of the intuitions behind the two ideas
followed by the model details.

\subsection{Intuitions\label{sec:intu}}
\textbf{Interpreting feedback in context}: The feedback is expected to link to all the other elements of the 
interaction~(Figure~\ref{fig:example}). The feedback is provided in the context
of the \textit{explanation} of the initial parse, as a proxy to the parse itself. As such, the feedback tends to use the same terminology as the explanation. For example, the SQL explanations of~\cite{Elgohary2020Speak}
express ``group by'' in simple language ``for each vote\_id, find ...''.
As a result, human-provided feedback never uses ``group by''.
We also notice that in several \splash\ examples, the feedback refers to particular steps in the explanation as in the examples in Figure~\ref{fig:example}. 
Unlike existing models~\cite{Elgohary2020Speak},
we replace the initial parse with its natural language explanation. 
Additionally, the feedback usually refers to columns/tables in the schema,
and could often be ambiguous when examined in isolation. Such ambiguities can be usually resolved by relying on the context provided by the question. For example, ``find last name'' in Figure~\ref{fig:example}
is interpreted as ``find last name besides first name'' rather than
``replace first name with last name'' because the question asks for 
the ``full name''. Our first key idea 
is based on grounding the elements of the interaction by combining
self-learned relations by transformer models \cite{vaswani2017attention} and
hard-coded relations that we define according to the possible ways
different elements can link to each other.

\textbf{Feedback describes a set of edits}:
The difference between the erroneous parse and the correct one can mostly be described as a few edits that need to be applied to the initial parse to correct 
its errors~(Section~\ref{sec:analysis}). Also, the feedback often only describes the edits to be made~\cite{Elgohary2020Speak}. As such, we can pose the task of correction with NL feedback as a semantic parsing task where we convert a natural language deception of the edits to a canonical form that can be applied deterministically to the initial parse to generate the corrected one. We train our model to generate SQL Edits~(Section~\ref{sec:sqledits})
rather than SQL queries.

\begin{figure}[!t]
\centering
  \includegraphics[width=\columnwidth]{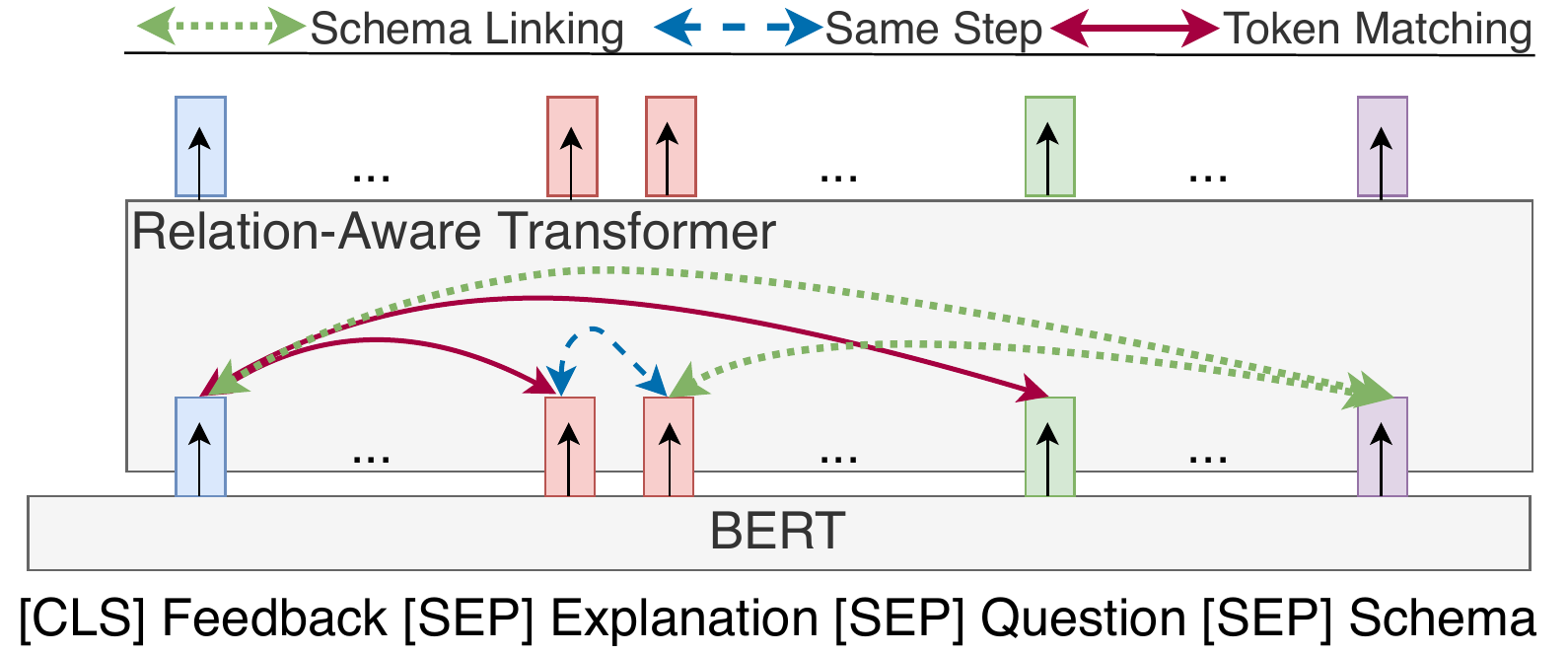}
    \caption{The Encoder of~\modelname~grounds the feedback into
    the explanation, the question, and the schema by (1) passing
    the concatenation of their tokens through BERT, then
    (2) combining self-learned and hard-coded relations in a relation-aware
    transformer. Three types of relations (Interaction Relations) link the
    individual tokens of the inputs. Question-Schema and Schema-Schema relations are not shown.}
  \label{fig:encoder}
 \end{figure}

\subsection{Encoder\label{sec:encoder}}
Our encoder~(Figure~\ref{fig:encoder}) starts with passing the concatenation of the
feedback, explanation, question, and schema through BERT~\cite{devlin2019bert}.
Following~\cite{rat-sql-2020,suhr-etal-2018-learning, Scholak2020Duorat}, we tokenize the column/table names and concatenate
them in one sequence (Schema) starting with the tokens of the tables followed by the tokens
of the columns. Then, we average the BERT embeddings of the tokens corresponding to each
column (table) to obtain one representation for the column (table).

\newcite{rat-sql-2020} study the text-to-SQL problem using the \spider\ dataset and show the benefit of injecting preexisting relations within the schema (column exists in a table, primary-foreign key), and between the question and schema items (column and table names) by: (1) name linking: link a question token to a column/table if the token and the item name match and (2) value linking: link a question token to a column if the token appears as a value under that column. To  incorporate such relations in their model, they use the relation-aware self-attention formulation presented in ~\cite{shaw-etal-2018-self}.
The relation-aware transformer ~\cite{shaw-etal-2018-self}
 assigns a learned embedding for each relation type and combines such embeddings with the self-attention of the original transformer model~\cite{vaswani2017attention}:
If a preexisting relation $r$ holds between two tokens, the embedding of $r$ is added as 
a bias term to the self-attention computation between the two tokens.
 
In addition to those relations, we define a new set of relations that aim at contextualizing the feedback with respect to the other elements of the interaction in our setup: 
(1) \textit{\textbf{[Feedback-Schema]}} We link the feedback to the schema the same way the question is linked to the schema via both name and value linking, (2)  \textit{\textbf{[Explanation-Schema]}} Columns and tables are mentioned with their exact
names in the explanation. We link the explanation to the schema only through exact 
name matching, (3)  \textit{\textbf{[Feedback-Question]}} We use partial~(at the lemma level)~and exact matching to 
link tokens in the feedback and the question, (4)  \textit{\textbf{[Feedback-Explanation]}} We link tokens in the feedback
to tokens in the explanation through partial and exact token matching. 
Since the feedback often refers to particular steps,
we link the feedback tokens to explanation tokens that occur in steps that are 
referred to in the feedback with a separate relation type that indicates
step reference in the feedback, and (5) \textit{\textbf{[Explanation-Explanation]}} We link explanation tokens that
occur within the same step. We use the same formulation of  relation-aware self-attention as ~\cite{rat-sql-2020} and add the relation-aware layers on top of BERT 
to integrate all relations into the model~(Figure~\ref{fig:encoder}).

\subsection{Decoder}
Using a standard teacher-forced cross-entropy loss, 
we train our model to generate \textit{linearized} SQL Edits~(Figure~\ref{fig:queryview}).
At training time, we compute the reference SQL Edit 
$\mathcal{D}_{\Tilde{\mathcal{P}}\rightarrow \hat{\mathcal{P}}}$ 
of the initial parse $\Tilde{\mathcal{P}}$ and the gold parse $\hat{\mathcal{P}}$\ (Section~\ref{sec:sqledits}).
Then we linearize $\mathcal{D}_{\Tilde{\mathcal{P}}\rightarrow \hat{\mathcal{P}}}$ by
listing the clause edits in a fixed order~(\texttt{FROM}, \texttt{WHERE},
\texttt{GROUP-BY}, ... etc.). The argument of  each clause---representing
one add or remove operation---is formatted as 
\verb!<CLAUSE> ADD/REMOVE ARG </CLAUSE>!.
We express SQL operators in~\verb!ARG! with natural  language
explanation as in \cite{Elgohary2020Speak}. For example, 
the argument \texttt{``AVG(grade)''} is expressed as ``average grade''. 
At inference time, we generate a corrected parse $\bar{\mathcal{P}}$
by applying the produced edit to the initial parse $\tilde{\mathcal{P}}$.

We use a standard transformer decoder that either generates tokens from the 
output vocab or copies columns and tables from the encoder output. 
Since all editing operations should be directed by the feedback, we 
tried splitting the attention to the encoder into two phases: First, we
attend to the feedback only and update the decoder state accordingly. 
Then, we use the updated decoder state to attend to the other 
inputs. With that, we only observed a marginal improvement of 0.5\% in the accuracy. 
We conduct all our experiments with standard decoder-encoder attention and plan to investigate other attention patterns in the future.

\section{Synthetic Feedback\label{sec:synthetic}}

\begin{table}[t]
	\centering
	\small
	\begin{tabular}{l}
          \toprule
		\textbf{Replace-Select-Column:} \\
		\ \ \ \ - replace \{NEW-COL\} with \{OLD-COL\}   \\
		\ \ \ \ - you should find \{OLD-COL\} instead \\
		\textbf{Add-Where-Condition:} \\
		\ \ \ \ - delete \{COL\} \{OPERATOR\} \{VALUE\} \\
		\textbf{Remove-Limit:} \\
		\ \ \ \ - only top \{LIMIT-VALUE\} rows are needed\\
		\bottomrule
	\end{tabular}
    \caption{Example SQL Editors with corresponding feedback templates.
    The synthesized feedback is reversing the edit applied to a correct SQL as
    our synthesis process starts with the gold SQL and reaches an initial SQL
    after applying the edit.}
    \label{tab:example_editors}
\end{table}

In this section, we describe our process for automatically synthesizing 
additional examples for training the correction model. 
Recall that each example consists of a question about a given schema paired with 
a gold parse, an initial erroneous parse, and  feedback. 
Starting with a seed of questions and their corresponding gold parses from 
~\spider's training set~(8,099 pairs)\footnote{We ensure there is no overlap between examples in the seed and the dev set of \splash.}, our synthesis process applies 
a sequence of SQL editing operations to the gold parse to reach an 
altered parse that we use as the initial parse~(Algorithm~\ref{alg:synthesis}).

By manually inspecting the edits~(Section~\ref{sec:sqledits}) we induce for the 
initial and gold parses in \splash\ training set, we define 26 SQL editors
and pair each editor with their most frequent corresponding feedback
template(s) (Examples in Table~\ref{tab:example_editors}).
We also associate each editor with a set of constraints that determines
whether it can be applied to a given SQL query~(e.g., the ``Remove-Limit'' 
editor can only be applied to a query that has a limit clause).

Algorithm~\ref{alg:synthesis} summarizes the synthesis process.
We start by creating $N$ (controls the size of the dataset) clones of each seed example.
\newcite{Elgohary2020Speak}'s analysis of \splash\ shows that
multiple mistakes might be present in the initial SQL, hence we allow our synthesis process to introduce up to four edits~(randomly decided in line:4) to each clone $p$.
For each editing step, we sample a feasible edit for the current parse~(line:5) 
with manually set probabilities for each edit to balance the number of times each
editor is applied in the final dataset. 
Applying an edit ~(line:6)~involves sampling
columns/tables from the current parse and/or the schema, sampling operators 
and values for altering conditions, and populating the corresponding
feedback template. We combine the feedback of all the applied editors  into one string and use it as the feedback of the synthesized example.

\section{Experiments\label{sec:exprs}}
\begin{table*}[t]
	\centering
	\small
	\begin{tabular}{l c |  c c c }
          \toprule
          & \textbf{Correction Acc.~(\%)} & \textbf{Edit $\downarrow$~(\%)} & \textbf{Edit $\uparrow$~(\%)} & \textbf{ Progress~(\%)} \\
	\hline
		Rule-based Re-ranking & 16.63 &  38.35 & 32.81 & -15.67 \\
		EditSQL+Feedback & 25.16  &  47.44 & 23.51 & 7.71 \\
		\modelname~(Ours) & \textbf{41.17} &  \textbf{72.41} & \textbf{16.93} & \textbf{36.99} \\
		\midrule
		Oracle Re-ranking& 36.38  & 34.69 & 1.04 & 31.22 \\
                \bottomrule
	\end{tabular}
    \caption{Comparing \modelname\ to \textbf{baselines} in~\cite{Elgohary2020Speak}: Rule-based Re-ranking
    and EditSQL+Feedback and to the beam re-ranking upper-bound.
    \textbf{Edit~$\downarrow$}~(\textbf{Edit~$\uparrow$}) is the percentage of examples
    on which the number of edits/errors strictly decreased~(increased). \textbf{Progress}
    is the average relative reduction in the number of edits~(Section~\ref{sec:exprs}).
    \newcite{Elgohary2020Speak} estimate the upper-bound on the correction accuracy as 81.5\%.
    }
    \label{tab:main_results}
\end{table*}

\vskip -1.5em
\textbf{Setup:}
We conduct our experiments using \splash\ \cite{Elgohary2020Speak}~(Section~\ref{sec:background})
whose train, dev, and test sets are of sizes 7481, 871, and 962, respectively.
Using our feedback synthesis process~(Section~\ref{sec:synthetic}), 
we generate 50,000 additional synthetic training examples. In our preliminary experiments, We found that training the model on the synthetic dataset first then continuing
on \splash\ outperforms mixing the synthetic and real examples and training on both of them simultaneously.
We train the model on the synthetic examples for 20,000 steps and continue training on the real examples until reaching 100,000 steps in total. We choose the 
best checkpoint based on the development set accuracy.
We varied the number of training steps on the synthetic examples and 20,000 steps achieved the highest accuracy on the dev set. 

\begin{algorithm}[t]
\small
\caption{Training Data Synthesis }\label{alg:synthesis}
\begin{algorithmic}[1]
\For{seed in \spider\ training set}
\For{$p$ in CLONE(seed, $N$)}
\State feedback = []
\For{$i = 1:\text{RAND-NUM-EDITS()}$}
\State $e \gets$ RAND-FEASIBLE-EDIT($p$)
\State $p$.APPLY-EDIT($e$)
\State feedback.ADD($e$.FEEDBACK())
\EndFor
\State \textbf{output:} seed.DB, seed.Question, $p$, 
\State \hspace{1.3cm} feedback, seed.Gold-SQL
\EndFor
\EndFor
\end{algorithmic}
\end{algorithm}

We use BERT-base-uncased~\cite{devlin2019bert} in all our experiments.
We set the number of layers in the relational-aware transformer to eight ~\cite{rat-sql-2020} and the number of decoder layers to two. We train with batches of size 24.
We use the Adam optimizer~\cite{kingma2015adam} for training.
We freeze BERT parameters during the first 5,000 warm-up steps and update
the rest of the parameters with a linearly increasing learning rate 
from zero to $5\times 10^{-4}$. Then, we linearly decrease the learning
rates from $5\times 10^{-5}$ for BERT and $5\times 10^{-4}$ for the 
other parameters to zero.\footnote{The learning rate schedule 
is  only dependent on the step number regardless of whether we are
training on the synthetic data or \splash. We tried resetting the
learning rates back to their maximum values after switching
to \splash, but did not observe any improvement in accuracy.}
We use beam search with a beam of size 20 and take the top-ranked 
beam that results in a valid SQL after applying the inferred edit.

\textbf{Evaluation:}
We follow ~\cite{Elgohary2020Speak} and use the \textit{correction accuracy} as our main evaluation measure: each example in \splash\ test set contains an initial parse~$\Tilde{\mathcal{P}}$ and a gold parse~$\hat{\mathcal{P}}$. With a predicted~(corrected)
parse by a correction model~$\bar{\mathcal{P}}$, they compute the correction accuracy
using the exact-set-match~\cite{yu-etal-2018-spider} between $\bar{\mathcal{P}}$ and $\hat{\mathcal{P}}$
averaged over all test examples. 
While useful, correction accuracy also has limitations. It expects models to be able to fully correct an erroneous parse
with only one  utterance of feedback as such, it is defined in terms of the exact match
between the corrected and the gold parse. We find (Table~\ref{tab:main_results}) that 
in several cases, models were still able to make \textit{progress} by reducing
the number of errors as measured by the edit size (Section~\ref{sec:sqledits})
after correction.
As such, we define another set of metrics to measure partial progress. We report~(Edit~$\downarrow$ and Edit~$\uparrow$ in Table~\ref{tab:main_results})~the percentage of examples on which 
the size of the edit set strictly decreased/increased. To combine 
Edit~$\downarrow$ and Edit~$\uparrow$ in one measure and account for the relative
reduction~(increase) in the number of edits, we define 
\[
\text{Progress} (\mathcal{S}) = \frac{1}{|S|}\sum_{\Tilde{\mathcal{P}}, \bar{\mathcal{P}}, \hat{\mathcal{P}} \in \mathcal{S}} 
\frac{|\mathcal{D}_{\Tilde{\mathcal{P}}\rightarrow \hat{\mathcal{P}}}| - |\mathcal{D}_{\bar{\mathcal{P}}\rightarrow \hat{\mathcal{P}}}|}
{|\mathcal{D}_{\Tilde{\mathcal{P}}\rightarrow \hat{\mathcal{P}}}|}.
\]
Given a test set $\mathcal{S}$, the Progress of a correction model is computed as 
the average relative edit reduction between the initial parse~$\Tilde{\mathcal{P}}$
and the gold parse $\hat{\mathcal{P}}$ by predicting a correction $\bar{\mathcal{P}}$ of
$\Tilde{\mathcal{P}}$. A perfect model that can fully correct all errors in the
initial parse would achieve a 100\% progress. A model can have a negative
progress~(e.g., Rule-based re-ranking in~Table~\ref{tab:main_results})
when it frequently predicts corrections with more errors than those in the initial parse.
Unlike correction accuracy, Progress is more aligned with 
user experience in an interactive environment~\cite{su2018natural} as
it assigns partial credit for fixing a subset of the errors and also, it penalizes
models that predict an even more erroneous parse after receiving feedback.

\textbf{Results:}
We compare~(Table~\ref{tab:main_results}) \modelname\ to the 
two top-performing baselines in
~\cite{Elgohary2020Speak} and also to the beam re-ranking upper-bound they report.
\modelname\ significantly increases the correction accuracy over the top baseline~(EditSQL+Feedback)
by more than 16\% and it also outperforms oracle re-ranking by around 5\%.
We also note that in 72.4\% of the test examples, \modelname\ was able to 
strictly reduce the number of errors in the initial parse~(Edit~$\downarrow$)
which potentially indicates a more positive user experience than the other
models. \modelname\ achieves 37\% Progress which indicates
faster convergence to the fully corrected parse than all the other models.

\section{Analysis\label{sec:analysis}}
\begin{figure*}[t!]
     \centering
     \begin{subfigure}[b]{0.24\textwidth}
         \centering
         \includegraphics[width=\textwidth]{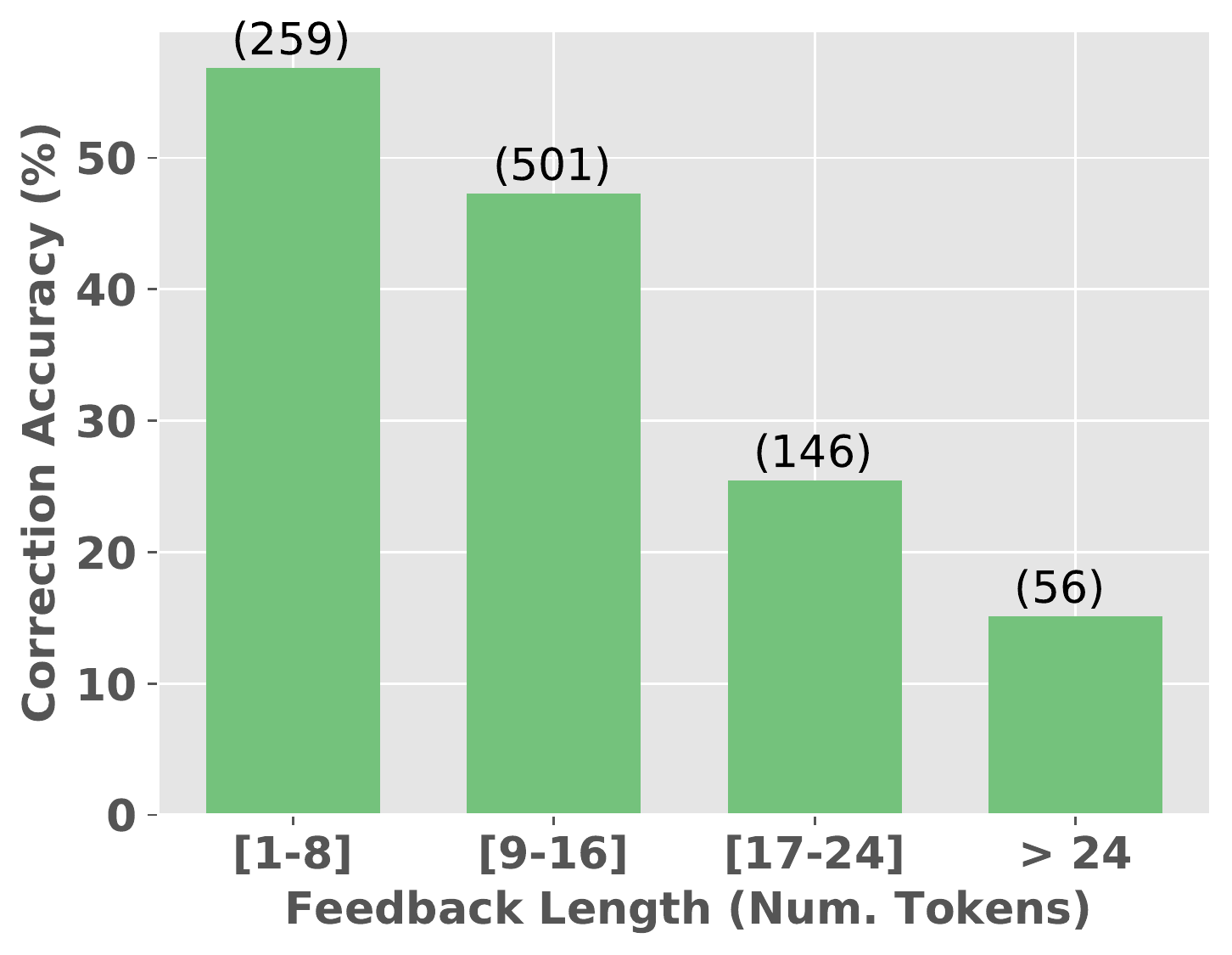}
         \caption{Feedback Length}
         \label{fig:acc_vs_flen}
     \end{subfigure}
     \hfill
     \begin{subfigure}[b]{0.24\textwidth}
         \centering
         \includegraphics[width=\textwidth]{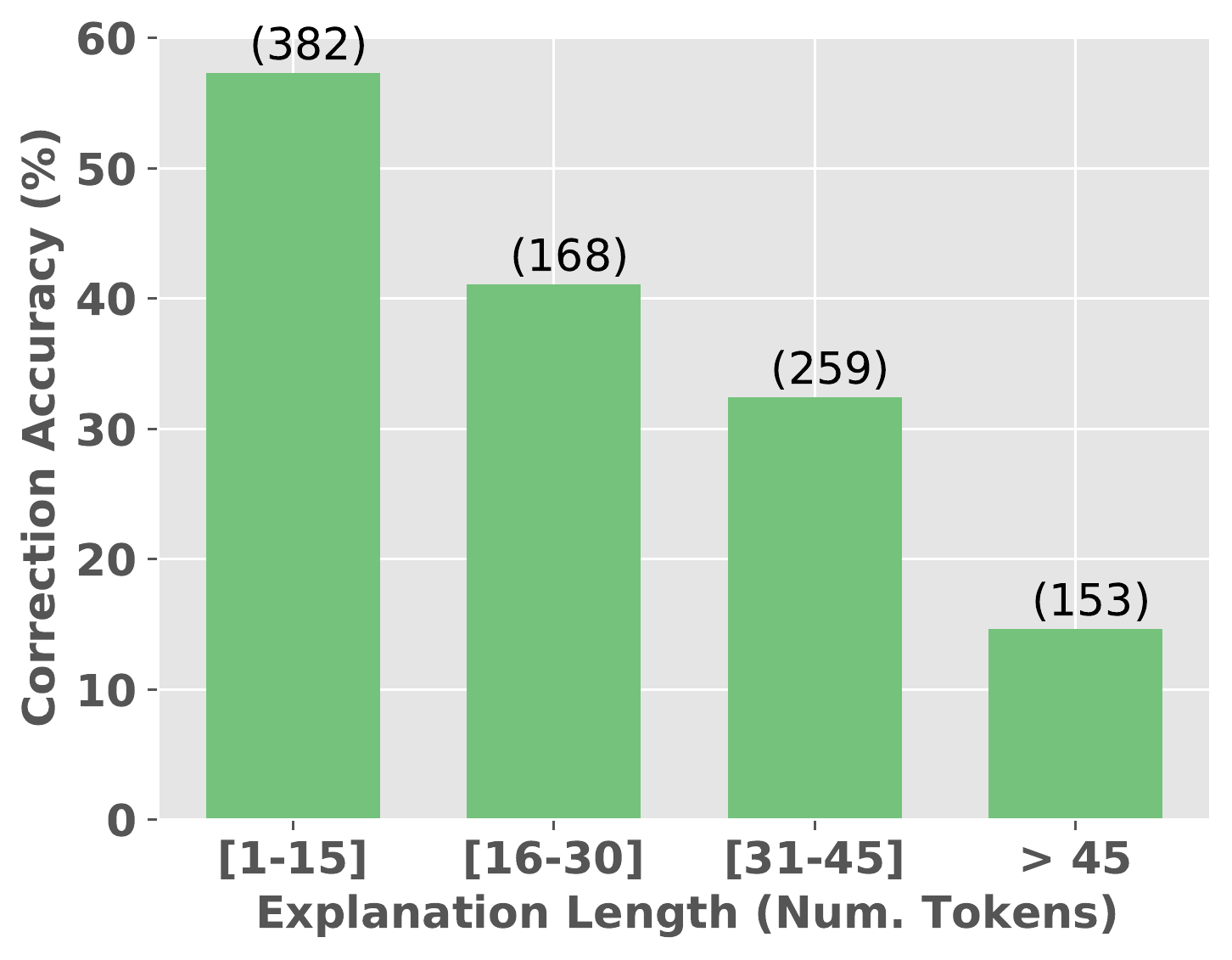}
         \caption{Explanation Length}
         \label{fig:acc_vs_elen}
     \end{subfigure}
     \hfill
     \begin{subfigure}[b]{0.24\textwidth}
         \centering
         \includegraphics[width=\textwidth]{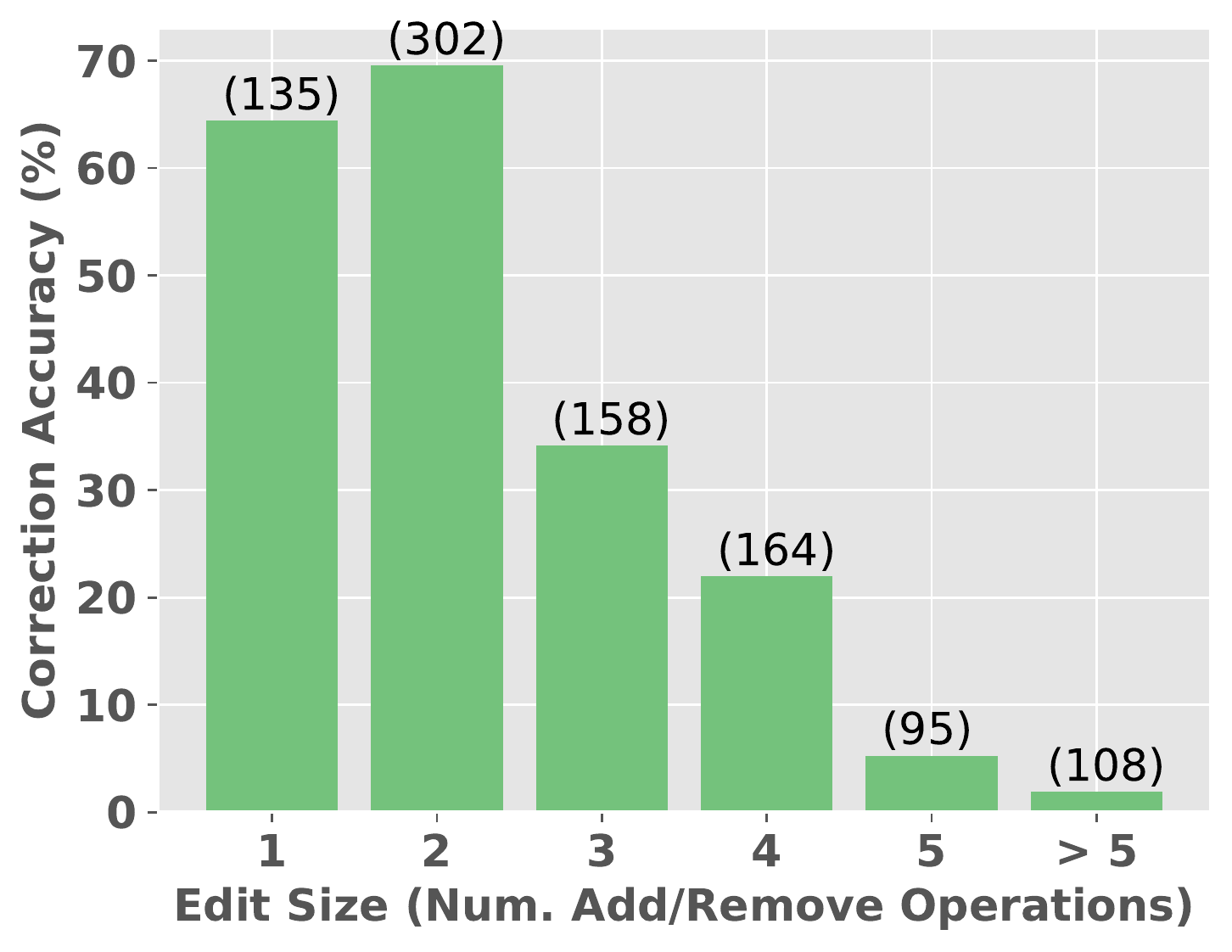}
         \caption{Reference Edit Size}
         \label{fig:acc_vs_editsize}
     \end{subfigure}
     \begin{subfigure}[b]{0.24\textwidth}
         \centering
         \includegraphics[width=\textwidth]{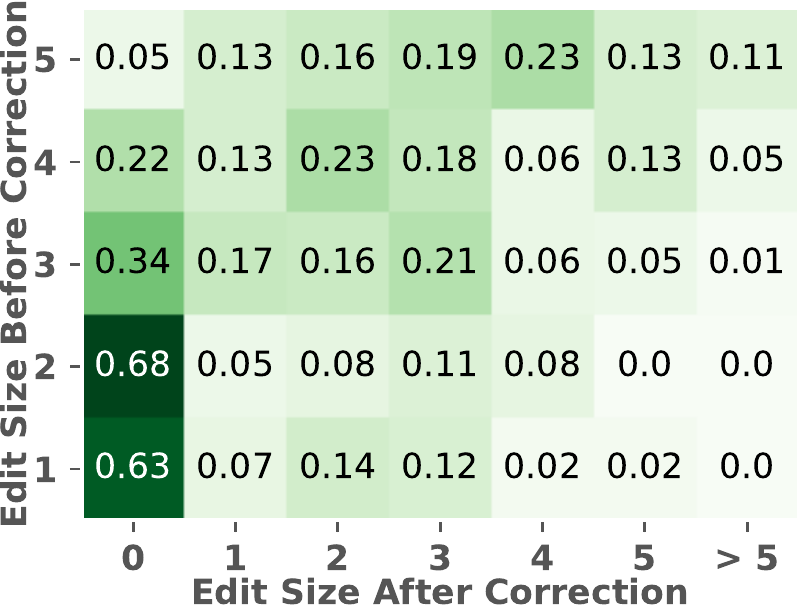}
         \caption{Transition of Edit Size}
         \label{fig:transition}
     \end{subfigure}     
        \caption{\textbf{a-c}: Breakdown of the correction accuracy on~\splash~test set
        by \textbf{(a)} feedback length, \textbf{(b)} explanation length,
        and \textbf{(c)} size of the reference edit (number of add or remove operations).
        The number of examples in each group is shown on top of the bars.
        \textbf{d:} Transitions in edit size after correction. For each edit size 
        of the initial parse (rows), we show the distribution of the edit size after
        correction.
        }
        \label{fig:breakdowns}
\end{figure*}

\subsection{Ablations}
Following the same experimental setup in Section~\ref{sec:exprs}, 
we compare \modelname\ to other variants with one ablated component at a time~(Table~\ref{tab:ablation}).
We ablate the \textbf{feedback}, the \textbf{explanation}, and the \textbf{question} from the encoder input. We also ablate the \textbf{interaction relations}~(Section~\ref{sec:encoder}) 
that we incorporate in the relation-aware transformer module. We 
only ablate the new relations we introduce to model the interaction~(shown in Figure~\ref{fig:encoder}),
but we keep the Question-Schema and Schema-Schema relations introduced 
in~\cite{rat-sql-2020}. For each such variant, we train for 20,000 steps on 
the synthetic dataset then continue training on~\splash\ until step 100,000.
We also train an ablated variant that does not use the \textbf{synthetic feedback} 
where we train for 100,000 steps only on~\splash. For all variants, we choose the 
checkpoint with the largest correction accuracy on the dev set and report the accuracy on the
~\splash\ test set.

The results in~Table~\ref{tab:ablation}~confirm the effectiveness of each
 component in our model. We find that the model is able to
 correct 19.8\% of the examples without
the feedback. We noticed that the ablated-feedback model almost reaches that accuracy only after
training on the synthetic data with very minor improvement (< 1\%) after
training on~\splash. Only using the question and the explanation, 
the model is able to learn about a set of systematic errors that parsers make and 
how they can be corrected~\cite{gupta2017deepfix, yin19acl}.

\begin{table}
\centering
\small
\begin{tabular}{lr}
\toprule
\modelname~& 41.17\\
$-$ Feedback & 19.81\\
$-$ Explanation & 26.80\\
$-$ Question & 38.27\\
$-$ Interaction Relations & 35.35\\
$-$ Synthetic Feedback & 35.01\\
 \bottomrule
\end{tabular}
\caption{Correction accuracy on \splash\ Test of \modelname\
versus variants with one ablated component each.
}
\label{tab:ablation}
\end{table}


\subsection{Error Analysis}

\begin{table}[t]
	\centering
	\small
	\begin{tabular}{p{7cm}}
          \toprule
		\textbf{\underline{Long Feedback Not Describing an Edit:}} \\
		``you should determine the major record format from the orchestra table and make sure it is arranged in ascending order of number of rows that appear for each major record format.''\\
		\midrule
        \textbf{\underline{Long Feedback Describing an Edit:}} \\
		``replace course id (both) with degree program id, first courses with student enrolment, course description with degree summary name, second courses with degree programs.''\\		
		\bottomrule
	\end{tabular}
    \caption{Example long feedback that~\modelname~struggles with. Top:
     The feedback describes a rewriting of the query rather than how to edit it.
     Bottom: The initial query has several errors and the feedback enumerates
     how to edit all of them.}
    \label{tab:long_feedback}
\end{table}

\begin{table}[t]
	\centering
	\small
	\begin{tabular}{p{7cm}}
          \toprule
		\textbf{\underline{Adding extra edits:}} \\
        \textbf{Ques.:} Which city and country is the Alton airport at?\\
        \textbf{Initial:}\hspace{3pt}\texttt{SELECT City, Country FROM airports WHERE AirportName = 'Alton' AND Country = 'USA'}\\
        \textbf{Feedback:} remove ``and country equals USA'' phrase.\\
         \textbf{Predicted:}\hspace{3pt}\texttt{<where> remove AirportName equals </where> \textcolor{red}{<where> remove Country equals </where>}}\\
        \textbf{Gold:}\hspace{3pt}\texttt{<where> remove AirportName equals </where>}\\
        \midrule
        \textbf{\underline{Failing to link feedback and schema:}} \\
        \textbf{Ques.:} What are the full names of all left handed players, in order of birth date?\\
        \textbf{Initial:}\hspace{3pt}\texttt{SELECT first\_name, last\_name FROM players ORDER BY birth\_date Asc}\\
        \textbf{Feedback:} make sure that player are left handed.\\
         \textbf{Predicted:} \texttt{<where> add \textcolor{red}{birth\_date} equals </where>}\\        
         \textbf{Gold:} \texttt{<where> add hand equals </where>}\\ 
    \bottomrule
	\end{tabular}
    \caption{Example failure cases of ~\modelname.
    }
    \label{tab:example_errors}
\end{table}

In Figure~\ref{fig:breakdowns},
we breakdown the correction accuracy by the feedback and explanation lengths (in number
of tokens) and by the reference edit size (number of required edit operations to 
fully correct the initial parse). The accuracy drops significantly when
the reference edit size exceeds two~(Figure~\ref{fig:acc_vs_editsize}), while
it declines more gradually as the feedback and explanation increase in length. 
We manually (Examples in Table~\ref{tab:long_feedback}) 
inspected the examples with longer feedback than 24, and found that 8\% of them the feedback is long because it describes how
to rewrite the whole query rather than being limited to only the edits to 
be made. In the remaining 92\%, the initial query had several errors (edit size of 5.5 on
average) with the corresponding feedback enumerating all of them.

Figure~\ref{fig:transition} shows how the number of errors (measured in edit size)
changes after correction. The figure shows that even for examples with a large
number of errors (four and five), the model is still able to reduce the number of errors
in most cases. We manually inspected the examples with only one error
that the model failed to correct. We found 15\% of them have either wrong or non-editing
feedback and in 29\% the model produced the correct edit but with additional irrelevant 
ones. The dominant source of error in the remaining examples is because of failures with linking
the feedback to the schema~(Examples in Table~\ref{tab:example_errors}).

\subsection{Cross-Parser Generalization}

\begin{figure}[!t]
\centering
  \includegraphics[width=0.8\columnwidth]{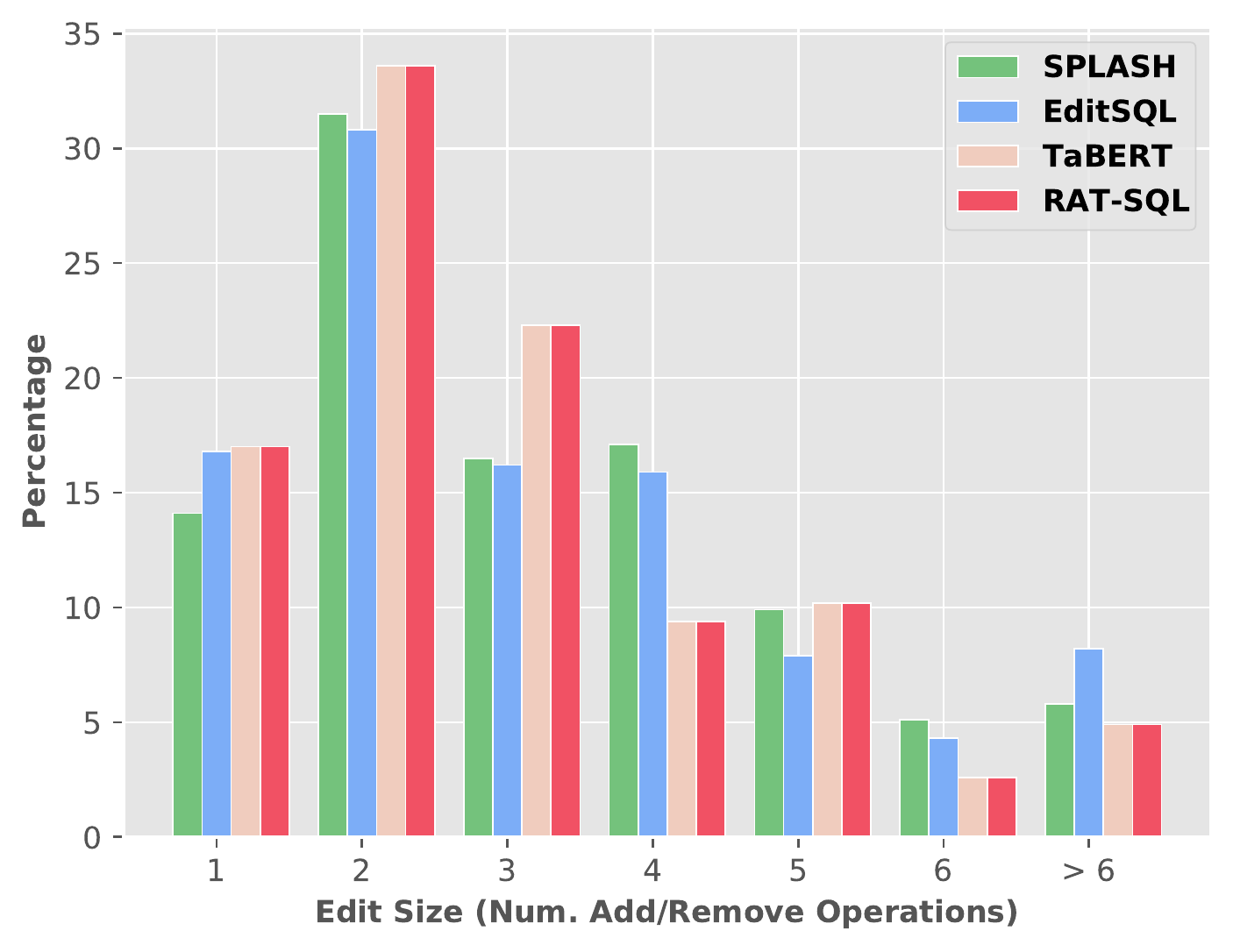}
    \caption{Distribution of Edit Size per example in \splash\ compared to the
    generalization test sets constructed based on EditSQL, TaBERT, and RAT-SQL.
    }
  \label{fig:editsize}
 \end{figure}

\begin{table*}[t!]
\centering
\small
\begin{tabular}{lc|ccc}
\toprule
& \textbf{Seq2Struct~(\splash)} & \textbf{EditSQL} & \textbf{TaBERT} & \textbf{RAT-SQL} \\
\hline
\multicolumn{3}{l}{\textbf{Correction Test Sets Summary}}\\
\midrule
Number of Examples & 962 & 330 & 267 & 208 \\
Average Feedback Length & 13.1 & 13.5 & 12.9 & 12.2 \\
Average Explanation Length & 26.4 & 28.3 & 32.2.9 & 34.0 \\
\midrule
\midrule
\multicolumn{3}{l}{\textbf{Semantic Parsing Accuracy (\%)}}\\
\midrule
Error Correction & 41.1 & 28.0 & 22.7 & 21.3 \\
No Interaction & \underline{41.3} & \underline{57.6} & \underline{65.2} & \underline{69.7} \\
End-to-End & \underline{61.6} & \underline{66.6} & \underline{71.1} & \underline{74.0} \\
$\Delta$ w/ Interaction  & \textbf{+20.3} & \textbf{+8.9} & \textbf{+5.9} & \textbf{+4.3} \\
 \bottomrule
\end{tabular}
\caption{Evaluating the zero-shot generalization of \modelname\ to different parsers
(EditSQL, TaBERT, and RAT-SQL) after training on~\splash\ that is
constructed based on the Seq2Struct parser. Top: Summary of the dataset
constructed based on each parser. Feedback and explanation length is the 
number of tokens. Bottom: The \textbf{Error Correction} accuracy on each test set
and the end-to-end accuracy of each parser \underline{on the full~\spider\ dev set} with and without interaction.
\textbf{$\mathbf{\Delta}$ w/ Interaction} is the gain in end-to-end accuracy with the interaction added.
}
\label{tab:gen_results}
\end{table*}

So far, we have been using~\splash\ for both training and testing.
The erroneous parses (and corresponding feedback) in
~\splash\ are based on the Seq2Struct parser~\cite{shin2019encoding}.
Recent progress in model architectures~\cite{rat-sql-2020} and
pre-training~\cite{yin-etal-2020-tabert, yu2020grappa} has led to parsers
that already outperform Seq2Struct by more than 30\% in
parsing accuracy.\footnote{https://yale-lily.github.io/spider}
 Here, we ask whether~\modelname\ that we train on \splash\  
 (and synthetic feedback) can generalize to parsing errors made by
more recent parsers without additional parser-specific training data.

We follow the same  crowdsourcing process used to construct
\splash\ (Section~\ref{sec:background}) to collect
three new test sets based on three recent text-to-SQL parsers: 
EditSQL~\cite{zhang19}, TaBERT~\cite{yin-etal-2020-tabert} and 
RAT-SQL~\cite{rat-sql-2020}.
Following~\newcite{Elgohary2020Speak},
we run each parser on~\spider\ dev set and only collect feedback for the examples with
incorrect parses that can be explained using their SQL explanation framework.
Table~\ref{tab:gen_results} (Top) summarizes the three new datasets
and compares them to~\splash\ test set. We note that the four datasets are
based on the same set of questions and databases~(\spider~dev).

Table~\ref{tab:gen_results} (Bottom) compares the  parsing
accuracy (measure by exact query match~\cite{yu-etal-2018-spider}) of each
parser when used by itself~\textbf{(No Interaction)} to 
integrating it with~\modelname. 
We report both the accuracy on the examples provided to \modelname\ (\textbf{Error Correction})
 and the \textbf{End-to-End} accuracy on the full~\spider\ dev set. \modelname\ significantly boosts the
accuracy of all parsers, but with a notable drop in the gains as the
accuracy of the parser improves. To explain that, in Figure~\ref{fig:editsize}
we compare the distribution of reference edit size across the four datasets.
The figure does not show any significant differences in the distributions
that would lead to such a drop in accuracy gain. Likewise, the distributions
of the feedback lengths are very similar (the mean is shown in Table~\ref{tab:gen_results}).
As parsers improve in accuracy, they tend
to make most of their errors on complex SQL queries. Although the number of
errors with each query does not significantly change~(Figure~\ref{fig:editsize}),
we hypothesize that localizing the errors in a complex initial parse, 
with a long explanation (Table~\ref{tab:gen_results}), is
the main  generalization bottleneck that future work needs to address.

\section{Related Work and Discussion \label{sec:related}}

\paragraph{Natural language to SQL:} Natural language interfaces to databases have been an active field of study for many years ~\cite{woods1972lunar,warren-pereira-1982-efficient, Popescu:2003:TTN:604045.604070,li2014constructing}. The development of new large scale datasets, such as WikiSQL~\cite{Zhong2017} and \spider~\cite{yu-etal-2018-spider}, has reignited the interest in this area with several new models introduced recently~\cite{choi2020ryansql,rat-sql-2020,Scholak2020Duorat}. Another related line of work has focused on conversation semantic parsing, e.g. SParC~\cite{Yuetal19Sparc}, CoSQL~\cite{yu-etal-2019-cosql}, and SMCalFlow~\cite{Andreas2020Task}, where parsers aim at modeling utterance sequentially and in context of previous utterances. 

\paragraph{Interactive Semantic Parsing:} Several previous studies have looked at the problem of improving semantic parser with feedback or human interactions~\cite{Clarke2010DrivingSP,artzi-zettlemoyer-2013-weakly}. ~\newcite{yao-etal-2019-model} and ~\newcite{gur-etal-2018-dialsql} ask yes/no and multiple-choice questions and use the answers in generating the pars. 
~\newcite{Elgohary2020Speak} introduce \splash\ (Section~\ref{sec:background}), a dataset for correcting semantic parsing with natural language feedback. Using language as a medium for providing feedback enables the human to provide rich open-form feedback in their natural way of communication giving them control and flexibility specifying what is wrong and how it should be corrected. Our work uses \splash\ and proposes to pose the problem of semantic parse correction as a parser editing problem with natural language feedback input. This is also related to recent work on casting text generation (e.g. summarization, grammatical error correction, sentence splitting, etc.) as a text editing task~\cite{malmi-etal-2019-encode,panthaplackel-etal-2020-learning,stahlberg-kumar-2020-seq2edits} where target texts are reconstructed from inputs using several edit operations.

\paragraph{Semantic Parsing with Synthetic Data:}
Semantic parsing systems have frequently used synthesized data to alleviate the challenge of labeled data scarcity. In their semantic parser overnight work, \citet{wang2015parser} proposed a method for training semantic parsers quickly in a new domain using synthetic data. They generate logical forms and canonical utterances and then paraphrase the canonical utterances via crowd-sourcing.  Several other approaches have demonstrated the benefit of adopting this approach to train semantic parsers in low-resource settings ~\cite{Su2017NL, Zhong2017, Cheng2018BuildingAN,  xu-etal-2020-autoqa}. 
Most recently,  synthetic data was used to continue to pre-train language models for semantic parsing tasks \cite{herzig2020tapas,yu2020grappa,yu2021score}.
We build on this line work by showing that we can generate synthetic data automatically without human involvement to simulate edits between an erroneous parse and a correct one.

\section{Conclusions and Future Work\label{sec:conc}}
We introduced a model, a data augmentation method, and analysis tools for correcting semantic parse errors in text-to-SQL through natural language feedback. Compared to previous models,
our model improves the correction accuracy by 16\% and
 boosts the end-to-end  parsing accuracy by up to 
20\% with only one turn of feedback.
Our work creates several avenues for future work: (1) improving the model by
better modeling the interaction between the inputs and exploring different patterns for
decoder-encoder attention, (2) evaluating existing methods for training with synthetic data~(e.g., curriculum learning ~\cite{bengio2009curriculum}), (3) optimizing the correction model for better user experience using the
progress measure we introduce, and (4) using the SQL edits scheme  in 
other related tasks such as conversational 
text-to-SQL parsing.

\section*{Acknowledgments}
This work has benefited greatly from discussions with Xiang Deng, Alex Polozov, Tao Yu, and Guoqing Zheng. We thank Pengcheng Yin for sharing TaBERT predictions before the official code release. We are 
very grateful to our reviewers for their insightful feedback and suggestions.

\bibliography{abrvs,ahmedg}
\bibliographystyle{acl_natbib}

\end{document}